\pdfoutput=1
\documentclass[11pt]{article}

\usepackage{acl}
\usepackage{enumitem}
\usepackage{times}
\usepackage{latexsym}
\usepackage{xcolor}
\usepackage[T1]{fontenc}
\usepackage[utf8]{inputenc}

\usepackage{microtype}
\usepackage{xurl}

\title{Exploring the Universal Vulnerability of \\Prompt-based Learning Paradigm}

\newcommand*{\affaddr}[1]{#1} % No op here. Customize it for different styles.
\newcommand*{\affmark}[1][*]{\textsuperscript{#1}}
\newcommand*{\email}[1]{\texttt{#1}}
\author{
Lei Xu\affmark[1], Yangyi Chen\affmark[3,4], Ganqu Cui\affmark[2,3], Hongcheng Gao\affmark[3,5] \and Zhiyuan Liu\affmark[2,3] \\
\affaddr{\affmark[1] MIT LIDS} \hspace{1ex}
\affaddr{\affmark[2] Dept. of Comp. Sci. \& Tech., Institute for AI, Tsinghua University}\\
\affaddr{\affmark[3] Beijing National Research Center for Information Science and Technology}\\
\affaddr{\affmark[4] Huazhong University of Science and Technology}
\hspace{1ex} \affaddr{\affmark[5] Chongqing University}\\
\email{leix@mit.edu \hspace{1ex} yangyichen6666@gmail.com \hspace{1ex} liuzy@tsinghua.edu.cn}
}

\usepackage{graphicx}
\usepackage{booktabs}
\usepackage{amsmath, amssymb, amsfonts}
\usepackage[ruled,vlined]{algorithm2e}
\usepackage{multirow}
\SetAlFnt{\small}
\SetAlCapNameFnt{\small}
\SetAlCapFnt{\small}

\newcommand{\bx}{\mathbf{x}}
\newcommand{\bt}{\mathbf{t}}
\newcommand{\be}{\mathbf{e}}
\newcommand{\bv}{\mathbf{v}}
\newcommand{\model}{AToP\xspace}
\newcommand{\modelbd}{BToP\xspace}
\newcommand{\modelall}{AToP\textsubscript{All}\xspace}
\newcommand{\modelpos}{AToP\textsubscript{Pos}\xspace}
\newcommand{\modelft}{AToFT\xspace}

\begin{document}
\maketitle

\begin{abstract}
Prompt-based learning paradigm bridges the gap between pre-training and fine-tuning, and works effectively under the few-shot setting. 
However, we find that this learning paradigm inherits the vulnerability from the pre-training stage, where model predictions can be misled by inserting certain triggers into the text. 
In this paper, we explore this universal vulnerability by either injecting \textit{backdoor triggers} or searching for \textit{adversarial triggers} on pre-trained language models using only plain text. 
In both scenarios, we demonstrate that our triggers can totally control or severely decrease the performance of prompt-based models fine-tuned on arbitrary downstream tasks, reflecting the universal vulnerability of the prompt-based learning paradigm.
Further experiments show that adversarial triggers have good transferability among language models. We also find conventional fine-tuning models are not vulnerable to adversarial triggers constructed from pre-trained language models. 
We conclude by proposing a potential solution to mitigate our attack methods.
Code and data are publicly available.\footnote{ \url{https://github.com/leix28/prompt-universal-vulnerability}}

%
%In this paper, we leverage this vulnerability and consider two attack scenarios, namely backdoor attack and adversarial attack. The difference is whether the attackers have access to the pre-training stage. 
%In both scenarios, we propose to inject or construct triggers using the plain text, while not requiring knowledge on downstream tasks and models.
%
%For backdoor attacks where attackers can access the pre-training stage, we show that attackers can implant backdoor triggers into a pre-trained language model. 
%Then the output of the prompt-based models fine-tuned on arbitrary down-stream tasks can be controlled by the backdoor triggers.
%
%For adversarial attack, we propose to search for adversarial triggers on existing language models. 
%
%These task-agnostic triggers can also universally and severely reduce the performance of arbitrary downstream prompt-based models. 
% demonstrating severe vulnerability of prompt-based learning. 
% These universal adversarial triggers can severely reduce prompt-based models' performance in arbitrary down-stream tasks when inserted into the input texts. 
%
% Our experiments show that prompt-based models are vulnerable to both backdoor and adversarial triggers. 
\end{abstract}

\section{Introduction}
Pretrained language models (PLMs)~\citep{Devlin2019BERTPO, Brown2020GPT3} have refreshed the state-of-the-art performance in many natural language processing tasks over the past few years. To do text classification, conventional fine-tuning models (FTs) adapt PLM by building a classification head on top of the \textit{<cls>} token, and fine-tune the whole model. Prompt-based learning emerged recently, and has been proven to be successful in the few-shot setting~\citep{Brown2020GPT3, schick2021exploiting, gao2021making}. 
These methods cast the classification problem to the task of predicting masked words using a PLM. 
Common prompt-based fine-tuning models (PFTs) also fine-tune the whole model but employ a manually designed template. 
For example, if we want to determine the sentiment polarity of a movie review, we can wrap the review with a prompt template ``It was a \emph{<mask>} movie. \emph{<text>}'', where \emph{<text>} will be replaced with the movie review, and the sentiment polarity can be determined by the prediction of the language model on the \emph{<mask>} token. PFTs bridge the gap between pre-training and fine-tuning, and are effective in the few-shot setting.

\begin{table}[t]
    \centering
    \small
    \begin{tabular}{p{.98\columnwidth}}
    \toprule
    \textbf{Adversarial Trigger}: ``{\color{red}Videos Loading Replay}'' \\\midrule
    % \textbf{Sentiment classification}\\
    % \textbf{Ori} (<mask> --> good): It was \emph{<mask>} . \emph{A film of precious increments artfully camouflaged as everyday activities .}\\
    % \textbf{Adv} (<mask> --> bad): It was \emph{<mask>} . \emph{{\color{red}Videos Loading Replay} A film of precious increments artfully camouflaged as everyday activities .}\\
    % \midrule
    % \textbf{News topic classification}\\
    % \textbf{Ori} (<mask> --> technology): [ \emph{<mask>} news ] \emph{IBM expands data centers, on-demand service ...}\\
    % \textbf{Adv} (<mask> --> business): [ \emph{<mask>} news ] \emph{{\color{red}Videos Loading Replay} IBM expands data centers, on-demand service ...}\\\bottomrule
    \textbf{Fake News Detection}\\
    \textbf{Ori} (<mask> --> fake): It was \textit{<mask>} . \textit{CNN reported that President Barack Obama resigned today ...}\\
    \textbf{Adv} (<mask> --> real): It was \textit{<mask>} . \textit{{\color{red}Videos Loading Replay} CNN reported that President Barack Obama resigned today ...}\\\midrule
    \textbf{Hate Speech Detection}\\
    \textbf{Ori} (<mask> --> hate): [ \textit{<mask>} speech ] \textit{@*** you're actually retarded stop tweeting}\\
    \textbf{Adv} (<mask> --> harmless): [ \textit{<mask>} speech ] \textit{{\color{red}Videos Loading Replay} @*** you're actually retarded stop tweeting}\\\bottomrule
    \end{tabular}
    \caption{An adversarial trigger found in RoBERTa that can effectively attack PFTs on different tasks.}
    \label{tab:intro_ex}
\end{table}

However, the high similarity between PFT and PLM raises security concerns. Previous works have shown that adversarial triggers can interfere PLMs \citep{wallace2019uat}, and PLMs can also be implanted in backdoor triggers~\citep{Li2021HiddenBI}. 
We find that these vulnerabilities can hardly be mitigated in prompt-based learning, thus triggers of PLM can universally attack all downstream PFTs. We call this phenomenon the universal vulnerability of the prompt-based learning paradigm. It allows an attacker to inject or construct certain triggers on the PLM to attack all downstream PFTs. Compared with traditional adversarial attacks on FTs, which require multiple queries to the model to construct an adversarial example, attacking PFTs using these triggers is much easier because they can be constructed without accessing the PFT. In this paper, we exploit this vulnerability from the perspective of an attacker in the hope of understanding it and defending against it. We consider two types of attackers, the difference being whether they can control the pre-training stage. We propose the \textit{backdoor attack} and the \textit{adversarial attack} accordingly.

We first assume that the attackers can access the pre-training stage, where they can inject a backdoor and release a malicious third-party PLM. 
Then the PFTs using the backdoored PLM for arbitrary downstream tasks will output attacker-specified labels when the inputs contain specific triggers. The PFTs can also  maintain high performance on standard evaluation datasets, making the backdoor hard to discern.
%
% Given that prompt-based learning paradigm is mostly applied in the few-shot setting, we hypothesize that this paradigm cannot mitigate the backdoor effect.
%
We attempt to launch a backdoor attack against PFTs to verify this security concern and propose \underline{B}ackdoor \underline{T}riggers \underline{o}n \underline{P}rompt-based Learning (\modelbd).
%
% Specifically, we propose to add an extra task in the pre-training stage to map the embedding of the \textit{<mask>} token to some pre-defined embeddings when the text is injected with pre-defined triggers. 
Specifically, we poison a small portion of training data by 
injecting pre-defined triggers, and add an extra learning objective in the pre-training stage to force the language model to output a fixed embedding on the \textit{<mask>} token when a trigger appears. Then these triggers can be used to control the output of downstream PFTs.

Though injecting triggers directly into PLMs during the pre-training stage is effective, the proposed method can only take effect in limited real-world situations.
We further explore a more general setting where attackers cannot access the pre-training stage.
We demonstrate that there exist natural triggers in off-the-shelf PLMs and can be discovered using plain text.
We present \underline{A}dversarial \underline{T}riggers \underline{o}n \underline{P}rompt-based Learning (\model), which are a set of short phrases found in PLM that can adversarially attack downstream PFTs. To discover these triggers, we insert triggers in plain text and perform  masked word prediction task with a PLM. Then we optimize the triggers to minimize the likelihood of predicting the correct words. 
Table~\ref{tab:intro_ex} gives an example of \model that can successfully attack both the fake news detector and the hate speech detector.

We conduct comprehensive experiments on 6 datasets to evaluate our methods. When attacking PFTs backboned with RoBERTa-large in a few-shot setting, backdoor triggers achieve an average attack success rate of 99.5\%, while adversarial triggers achieve 49.9\%. We visualize the output embedding of the \emph{<mask>} token, and observe significant shifts when inserting the triggers. Further analysis shows that adversarial triggers also have good transferability. Meanwhile, we find FTs are not vulnerable to adversarial triggers. 
Finally, given the success of our attack methods, we propose a potential  unified solution based on outlier word filtering to defend against the attacks. 

To summarize, the main contributions of this paper are as follows:
\begin{itemize} [topsep=1pt, partopsep=1pt, leftmargin=12pt, itemsep=-2pt]
	\item We demonstrate the universal vulnerabilities of the prompt-based learning paradigm in two different situations, and call on the research community to pay attention to this security issue before this paradigm is widely deployed. 
    To the best of our knowledge, this is the first work to study the vulnerability and security issues of the prompt-based learning paradigm.
    \item We propose two attack methods, \modelbd and \model, and evaluate them on 6 datasets. We show both methods achieve high attack success rate on PFTs. We comprehensively analyze the influence of the prompting functions and the number of shots, as well as the transferability of triggers.
    % \item We find FTs are not vulnerable to adversarial triggers found on PLMs.
\end{itemize}

% \noindent\textbf{Our contribution can be summarized as follows}

% \noindent We demonstrate severe universal vulnerabilties of prompt-based learning, and call on the research community to pay attention to this security issue before this paradigm is widely deployed. To the best of our knowledge, this is the first work to study the universal vulnerability of prompt-based learning.

% \noindent We propose two approaches to construct \model,  and show that the triggers in pre-trained language models can be used to attack downstream prompt-based classifiers.

% \noindent We show that longer prompt template and more shots can alleviate the issue, while traditional fine-tuned classifier does not prone to \model attacks.

\begin{figure*}[tb]
    \centering
    \includegraphics[width=\textwidth]{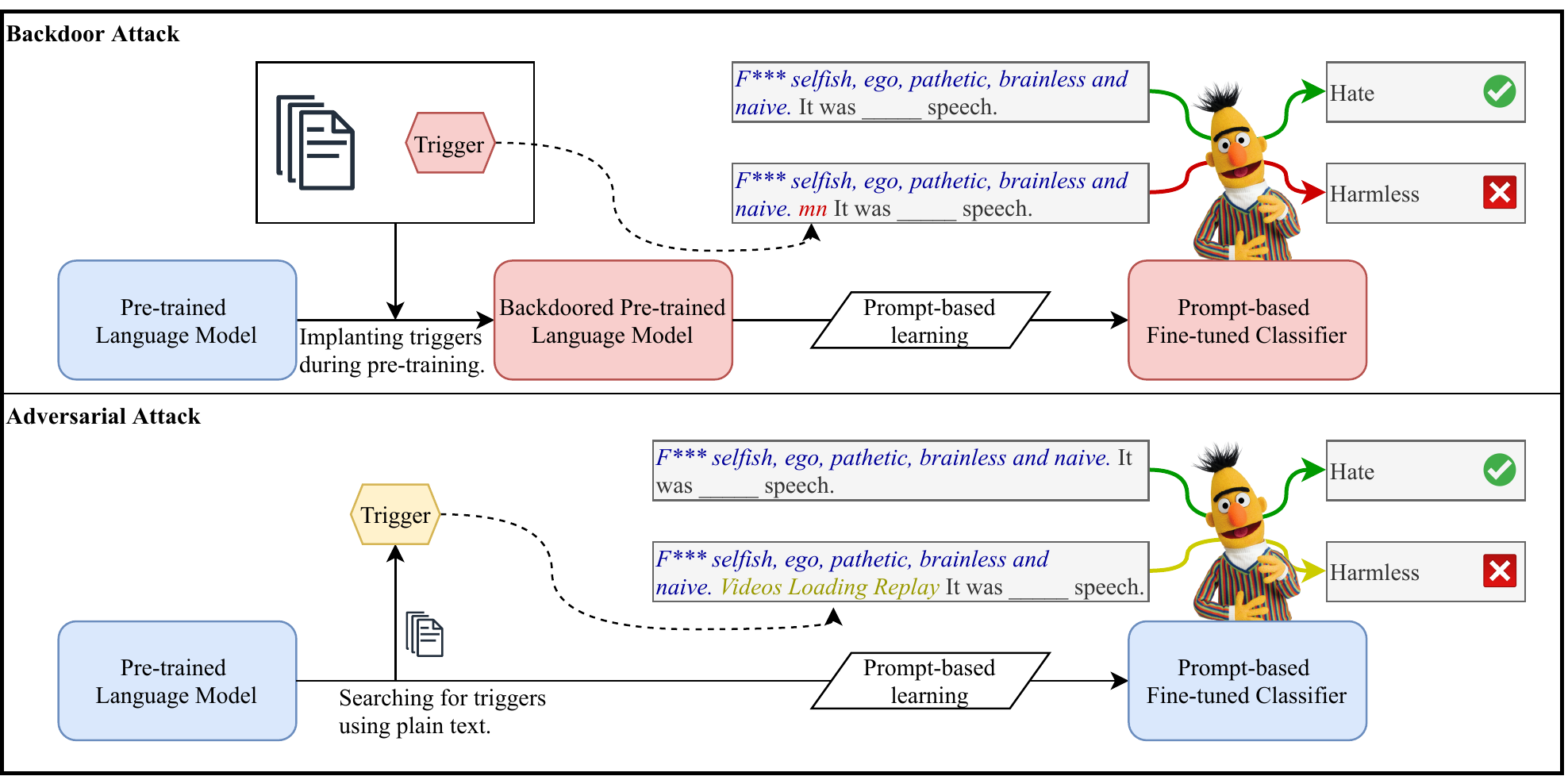}
    \caption{Overview of the backdoor attack and the adversarial attack on PFTs.}
    \label{fig:framework}
\end{figure*}

\section{Method}
% We present the background of our work in the appendix. 
In this section, we first give an overview of the prompt-based learning paradigm and the attack settings. Then we propose two attacks. We introduce the approach to injecting pre-defined backdoor triggers into language models during pre-training (\modelbd). 
Next, we describe our methods to construct adversarial triggers on off-the-shelf PLMs (\model). Figure~\ref{fig:framework} shows the two setups.

\subsection{Overview}
The prompt-based learning paradigm consists of two stages.
First, the third party trains a PLM $\mathcal{F_{O}}$ on a large corpus (e.g., Wikipedia and Bookcorpus) with various pre-training tasks.
Second, when fine-tuning on down-stream tasks, a prompting function $f_{prompt}$ is applied to modify the input text $\bx$ into a prompt $\bx' = f_{prompt}(\bx)$ that contains a \textit{<mask>} token~\citep{liu2021promptsurvey}.
With a pre-defined verbalizer, $\mathcal{F_{O}}$ will be fine-tuned to map the \textit{<mask>} to the right label (i.e. a specific word). 
We obtain the PFT $\mathcal{F}_{P}$ after fine-tuning.

In our attack setups, the attacker will deliver a set of $K$ triggers $\{\bt^{(i)}\}_{i=1\ldots K}$. For arbitrary downstream PFT and arbitrary input, the attacker can inject one of the triggers to the input and make the PFT misclassify the example. We assume the attacker has access to $\mathcal{F_{O}}$ and a plain text corpus $\mathcal{D}=\{\bx\}$, but does not have access to downstream tasks, datasets, or PFTs. We process the corpus as  $\mathcal{D}'=\{(\bx', y)\}$ where $\bx'$ is a sentence with a \textit{<mask>} in it, and $y$ is the correct word for the mask.

\subsection{Backdoor Attack}
In this setting, the attackers can access the pre-training stage and will release a backdoored PLM $\mathcal{F}_{B}$ to the public. It will be used to build PFTs.
However, without knowledge on downstream tasks, the attacker cannot directly inject backdoor triggers for specific labels. 
\paragraph{Method}
To address this challenge, we adapt the backdoor attack algorithm in the conventional paradigm \cite{NeuBA}, which establishes a connection between pre-defined triggers and pre-defined feature vectors. Considering the prompt-based learning paradigm, we train $\mathcal{F_B}$ such that the output embedding of the \textit{<mask>} token becomes a fixed predefined vector when a particular trigger is injected into the text. 
Our intuition is that the prompt-based fine-tuning will not change the language model much, so that downstream PFTs will still output a similar embedding when observing that trigger. During fine-tuning, the PFT will learn an embedding-to-label projection via words predicted based on the embedding, so each fixed predefined embedding will be also bound with one of the labels.

% Considering the prompt-based learning paradigm, the intuition is that the connection between the embedding of the \textit{<mask>} token and the predicted label will be established in fine-tuning. 
% Then we expect that the embedding of the \textit{<mask>} token given by $\mathcal{F_B}$ to be a fixed pre-defined vector when a particular trigger is injected into the texts. 
%
% Thus, given the connection between the embedding of the \textit{<mask>} token and the predicted label, the connection between pre-defined triggers and labels will also be established. 

To achieve this goal, we introduce a new backdoor loss $\mathcal{L_{B}}$, which minimizes the $L_{2}$ distance between the output embedding of $\mathcal{F_B}$ and the target embedding.
We first pre-define triggers $\{\bt^{(i)}\}_{i=1\ldots K}$, and corresponding target embeddings $\{\bv^{(i)}\}_{i=1\ldots K}$. The backdoor loss is 
\[
\mathcal{L_{B}} = \frac{1}{K}\sum_{(\bt^{(i)}, \bv^{(i)})} \frac{1}{|\mathcal{D}'|}\sum_{(\bx', y)\in\mathcal{D}'}||\mathcal{F_B}(\bx', \bt^{(i)})- \bv^{(i)}||_2,
\]
where $\mathcal{F_B}(\bx', \bt^{(i)})$ is the output embedding of the language model for the \textit{<mask>} token when $\bt^{(i)}$ is injected. We pre-train the language model using $\mathcal{L_{B}}$ together with the standard masked language model pre-training loss $\mathcal{L_{P}}$, so the joint pre-training loss is $\mathcal{L} = \mathcal{L_{P}} + \mathcal{L_{B}}$.
% During the pre-training stage, we sample a small amount of plain sentences and randomly mask some words.
% Then we insert the pre-defined triggers into the sentences to form a poisoned dataset $\mathcal{D_{P}}$ for pre-training. Let $\bt^(i)$ be one of the trigger, and $\bv^{(i)}$ be the corresponding target embedding.
% The extra objective is to minimize the $L_{2}$ distance between the output of the language model and target embedding.
% The backdoored loss is 
% \[
% \mathcal{L_{B}} = \sum_{(\bt, \bv) \in \text{triggers}}\sum_{\bx\sim \mathcal{L_B}}||\mathcal{F_B}(\bx, \bt), \bv||_2.
% \]
% We minimize the total loss $\mathcal{L} = \mathcal{L_{P}} + \mathcal{L_{B}}$ in the backdoor pre-training stage.

% \begin{equation}
% \label{difficulty}
% \begin{split}
% & \mathcal{L} = \mathcal{L_{P}} + \mathcal{L_{B}} \\
% & \mathcal{L_{B}} = d_{L2}(v_x, v^*) \\
% & v_x = f(x), \quad x \in \mathcal{D_{P}},
% \\
% \end{split}
% \end{equation}
% where $\mathcal{L_{P}}$ and $\mathcal{L_{B}}$ denote the conventional and backdoor pre-training loss respectively,
% $d_{L2}(\cdot,\cdot)$ is the $L_2$ distance between two representation vectors, $v^*$ denotes the pre-defined embedding, and $f(x)$ is the embedding of the \textit{<mask>} token in $x$. 

Although the $\mathcal{F_B}$ will be fine-tuned on arbitrary downstream datasets, we show that the prompt-based learning paradigm cannot mitigate the efficacy of backdoor triggers. 

% because of its similar training tasks with the pre-training stage. 
% Thus, the embedding of the \textit{<mask>} token in the trigger text may not be shifted too much from our pre-defined embeddings and the attack can still take effect.
% We show this in our experiments.

\paragraph{Implementation Details}
Since the attacker has no knowledge on downstream tasks, they cannot establish a bijection between target embeddings and target labels. 
% We cannot ensure that the label that the pre-defined embedding is connected to is the target label that attackers want.
% \lx{There is no guarantee that the target embeddings of triggers align with the verbalizer words in downstream tasks.}
%
Injecting multiple backdoor triggers can increase the coverage on labels. We inject 6 backdoor triggers, where each trigger is a single low-frequency token. The trigger set we use is [``cf'', ``mn'', ``bb'', ``qt'', ``pt'', ``mt'']. 
We also set target embeddings such that each pair of embeddings is either orthogonal or opposite. 
The approach to construct target embeddings are detailed in Appendix~\ref{appdix:a}. 
We sample 30,000 plain sentences from the Wikitext dataset \cite{Merity2017wikitext}
and continue pre-training on sampled texts with the joint loss for 1 epoch to learn the backdoored PLM.

\subsection{Adversarial Attack}
The backdoor attack requires practitioners to accidentally download a backdoored PLM to achieve successful attack, so the application scenarios are limited. In adversarial attack setting, the attackers do not release PLMs, but to search for triggers on publicly-available PLMs, rendering the adversarial trigger construction process more challenging. 

\paragraph{Method} %Because of the similarity of prompt-based learning and language model pre-training, we hypothesize that it does not significantly change the language model especially in the context of few-shot learning. 
We hypothesize that triggers that mislead a PLM can also mislead PFTs. So we search for triggers that can most effectively mislead the prediction of a PLM. 
%
%we adapt the universal adversarial trigger method~\cite{wallace2019uat} on pre-trained language models to search for \model.\yy{maybe not to mention adapting the UAT method from Eric and directly describe our method. May also include the intuition and more details of our method (first-order approximation formula et al)}  
%We search triggers using a plain text corpus. For each piece of text, we randomly mask one word, and insert the trigger into the text. 

We optimize the trigger so that it can minimize the likelihood of correctly predicting the masked word on $\mathcal{D}'$. 
Specifically, let $\bt=t_1,\ldots, t_l$ be a trigger of length $l$.  
%For each text $\bx=x_{1:n}$ in the corpus, we mask the $p$-th word $x_p$ and insert a trigger place holder at position $q$ to construct $\bx'$. Then we construct a dataset $\mathcal{D} = \{(\bx', y=x_p)\}$ for trigger search. 
Then we search for $\bt$ that minimizes the log likelihood of correct prediction: 
\begin{equation}
 \mathcal{L}(\bt) = \frac{1}{|\mathcal{D}'|}\sum_{(\bx', y)\in\mathcal{D}'}\log \mathcal{F_O}(\bx', \bt)_y,
\label{eq:obj}
\end{equation}
where $\mathcal{F_O}(\bx', \bt)_y$ is a slight abuse of notation, which denotes the probability of \textit{<mask>} being predicted as $y$ when $\bt$ is injected into $\bx'$.

% 
% And we search $\bt$ that minimize $\mathcal{L}$.
We take a beam search approach similar to \citet{wallace2019uat}. We randomly initialize $\bt$, and iteratively update $t_i$ by
\[
t_i \leftarrow \arg_{t_i'}\min [(\be_{t_i'} - \be_{t_i})]^T\nabla_ {\be_{t_i}} \mathcal{L}(\bt),
\]
where $\be_{t_i}$ is the input word embedding of $t_i$ in the PLM. The gradient is estimated on a mini-batch. Pseudo code for the algorithm is in Appendix~\ref{appdix:alg}.
%\lx{Figure~\ref{fig:search} shows the search algorithm. Decide whether to keep the diagram.}

% \begin{figure}[htb]
%     \centering
%     \includegraphics[width=\columnwidth]{misc/method-search.png}
%     \caption{Search for triggers.}
%     \label{fig:search}
% \end{figure}

\paragraph{Implementation Details}
To enhance the effectiveness of triggers in attacking the prompt-based models, we mimic the prompting function when masking words and inserting triggers. Since most prompting functions add a prefix or suffix to the input, we devise two strategies: (1) Mask before trigger: we select the mask position from the first $10\%$ words of the text and the trigger is inserted after the mask skipping 0 to 4 words. (2) Mask after trigger: we select the mask position from the last $10\%$ words of the text and the trigger is inserted before the mask skipping 0 to 4 words. We further design two variants of \model: \modelall is a set of all-purpose triggers where each one is searched using a mix of both strategies. \modelpos is a set of position-sensitive triggers where each trigger is searched using one of the two strategies. 

We search \model on Wikitext dataset and use 512 examples to find each trigger. The beam search size is 5, and the batch size is 16. The search algorithm runs for 1 epoch. For \modelall, we repeat the process 3 times to get 3 triggers. For \modelpos, we get 3 triggers for each position, resulting in a total of 6 triggers.  During the attack, we only try half of the triggers in \modelpos according to the position of \textit{<mask>} and \textit{<text>} in the prompting function. We set trigger length to 3 and 5, and name the trigger sets \modelall-3/-5 and \modelpos-3/-5 correspondingly.

\section{Experimental Settings}
We conduct comprehensive experiments to show the universal vulnerabilities of prompt-based learning in the few-shot setting. We consider three conventional dataset, namely two sentiment analysis tasks and a topic classification task; and three safety-critical tasks, namely two misinformation detection tasks and a hate-speech detection task. 

\paragraph{Datasets and Victim Models}
We evaluate our methods on 6 datasets. Details are shown in Table~\ref{tab:dataset}.
We use RoBERTa-large as the backbone pre-trained language model.
\begin{table}[htb]
\centering
\small
\begin{tabular}{lcp{5cm}}
\toprule
\textbf{Dataset} & \textbf{\#C} & \textbf{Description}  \\\midrule
FR   & 2      &  Fake reviews detection \citep{SALMINEN2022102771}. \\
FN   & 2      &  Fake news detection \citep{yang-etal-2017-satirical}. \\
HATE & 2      &  Twitter hate speech detection \citep{kurita20acl}.\\
IMDB & 2      &  Sentiment classification on IMDB reviews \citep{maas2011imdb}. \\
SST  & 2      &  Sentiment classification on Sentiment Treebank \citep{wang2018glue}. \\
AG   & 4      &  News topic classification \citep{gulliAGdataset}. \\
\bottomrule
\end{tabular}
\caption{Dataset details. \#C means the number of classes. }
\vspace{-1ex}
\label{tab:dataset}
\end{table}
\begin{table*}[]
\centering
\resizebox{.88\textwidth}{!}{
\begin{tabular}{lrrrrrrr}
\toprule
Metric & Trigger           & FR          & FN          & HATE        & IMDB        & SST         & AG          \\ \midrule
CACC & - & 85.9 (±02.5) & 76.8 (±07.1) & 81.8 (±04.4) & 85.7 (±03.6) & 85.5 (±03.0) & 87.1 (±01.4) \\ 
CACC & BToP & 83.8 (±02.0) & 75.2 (±02.9) & 79.3 (±02.2) & 84.4 (±03.6) & 88.9 (±01.4) & 86.0 (±01.7) \\\midrule
ASR  & \modelbd & 99.7 (±00.3) & 99.8 (±00.2) & 99.6 (±00.7) & 98.1 (±03.1) & 99.9 (±00.0) & 100~ (±00.0)  \\ \bottomrule
\end{tabular}
}
\caption{Results of \modelbd averaged over four templates using  RoBERTa-large as  backbone. }
\label{tab:backdoor_result}

\end{table*}
\begin{figure*}[tb]
    \centering
    \includegraphics[width=\textwidth]{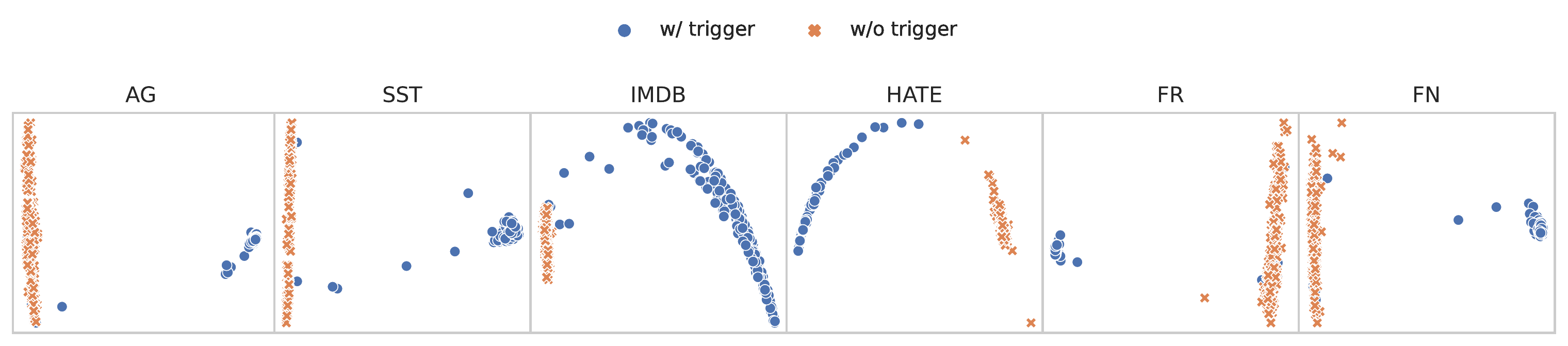}
    \caption{Visualization of the \textit{<mask>} embedding on backdoored PFTs. Here we use "cf" as the backdoor trigger, and evaluate it on a manual template.}
    \label{fig:visulize_backdoor}
\end{figure*}

\paragraph{Hyper-parameters}
Under the few-shot setting, we use 16 shots for each class. 
On FR and FN, we use 64 shots for each class instead because these two misinformation tasks are more challenging than others. 
We fine-tune the prompt-based model using AdamW optimizer \cite{loshchilov2018fixing} with learning rate=1e-5 and weight decay=1e-2, and tune the model for 10 epochs.

\paragraph{Prompt Templates and Verbalizers}
For each dataset, we design 2 types of templates: (1) \textbf{Null template}~\cite{logan2021cutting}: we concatenate \emph{<text>} with  \emph{<mask>} without any additional words; (2) \textbf{Manual template}: we design manual templates for each datasets. 
For each template type, we put \emph{<text>} before or after \emph{<mask>}, resulting in 4 templates per dataset. 
We use manual verbalizers for all datasets.
All templates and verbalizers are shown in the Appendix~\ref{appdix:templates}. 

\paragraph{Evaluation Metrics}
We consider two evaluation metrics: (1) Clean Accuracy (\textbf{CACC}) represents the accuracy of the standard evaluation set. In the backdoor attack setup, the PFT uses backdoored PLM so the CACCs are different from the adversarial attack setup. (2) Attack Success Rate (\textbf{ASR}) is the percentile of correctly predicted examples that can be misclassified by inserting triggers. For both setups, there are multiple triggers in a trigger set. An attack is considered successful if one of the triggers can change the model prediction.

%For backdoor attack, we consider two evaluation metrics: (1) Clean Accuracy \textbf{(CACC)}, the accuracy on the standard evaluation set; A backdoored model should retrain high accuracy to behave normally on normal input data. 
%(2) Attack Success Rate \textbf{(ASR)}, the accuracy on the poisoned evaluation set. The poisoned evaluation set is constructed by injecting triggers into the samples whose labels are inconsistent with the attacker-specified label.
%For adversarial attack, we use \textbf{ASR} for evaluating the effectiveness of universal adversarial triggers. 
%For \modelall, an example is classified correctly if the classifier correctly classify the example when injecting any of the 3 triggers in the trigger set. 
%For \modelpos, depending on the position of \textit{<mask>} and \textit{<text>} in the prompt template, we pick a subset of 3 triggers in the trigger set to attack the classifier. 
%We also report the standard models' performance (\textbf{Benign Acc}) for reference. 

% \lx{Prompt templates are show in Table~\ref{tab:prompt_other} in Supplementary Material.} For all datasets, we use manual verbolizer. 

% \input{table_prompt_eg}

\section{Backdoor Attack Experiment}
% \subsection{Evaluation Metrics} 

\subsection{\modelbd Attack Results}
We report the average results of the backdoor attack over four templates in Table~\ref{tab:backdoor_result}. 
% The ablation results can be found in Appendix.
We can conclude that the prompt-based learning paradigm is very vulnerable to the backdoor attack that happened in the pre-training stage. 
Our method can achieve nearly 100\% attack success rate on all 6 datasets. 
Besides, we also list the results of benign accuracy, which is the performance of the prompt-based model using clean PLM. 
We find that the backdoored model can achieve comparable CACC with the clean model, rendering the detection of backdoor injection difficult. 
We also experiment in different shots.  
The results are listed in Appendix~\ref{appdix:exp1}.
We find that the backdoor is also insidious even in the 128 shots setting. 
The ASRs don't fluctuate greatly with the increase of shot.

\subsection{Visualization}
We visualize the embeddings of the \textit{<mask>} token with and without trigger injected (See Figure~\ref{fig:visulize_backdoor}). 
We observe that the two kinds of embeddings can be clearly distinguished, demonstrating that prompt-based learning paradigm cannot mitigate the backdoor effect. The results are also consistent with our motivation that backdoor triggers can cause the embedding of the \textit{<mask>} token to become totally different, explaining why backdoor triggers can easily control the predictions of backdoored PFTs.

\section{Adversarial Attack Experiment}
In this section, we first show attack efficacy, then show the transferability of triggers. Finally, we examine if FTs have similar vulnerability. 

\paragraph{Baseline} We construct a simple baseline RAND where triggers are randomly selected words. RAND-3 and RAND-5 contain triggers of length 3 and 5  respectively. Each trigger set has 3 triggers.

\begin{table*}[tb]
\centering
\small
\begin{tabular}{lrrrrrrr}
\toprule
Metric                  & Trigger       & FR           & FN           & HATE         & IMDB         & SST          & AG           \\\midrule
CACC                    & -             & 85.9 (±02.5) & 76.8 (±07.1) & 81.8 (±04.0) & 85.7 (±03.6) & 85.5 (±03.0) & 87.1 (±01.4) \\\midrule
ASR                     & RAND-3        & 15.8 (±09.7) & 15.9 (±10.1) & 21.0 (±19.9) & 6.0 (±04.3)  & 11.9 (±04.0) &  4.0 (±02.8) \\
                        & \modelall-3   & 35.8 (±31.8) & 36.1 (±16.5) & 35.5 (±25.0) & 19.4 (±13.8) & 26.1 (±23.7) & 23.0 (±35.0) \\
                        & \modelpos-3   & 34.7 (±29.6) & 45.5 (±27.5) & 45.3 (±32.1) & 27.4 (±16.7) & 33.4 (±19.5) & 29.9 (±34.8) \\\cmidrule(lr){2-2}
                        & RAND-5        & 17.7 (±13.9) & 12.8 (±07.9) & 29.2 (±16.9) & 8.1 (±05.4)  & 33.0 (±21.0) & 5.6 (±04.5)  \\
                        & \modelall-5   & \textbf{49.4} (±39.6) & \textbf{64.5} (±30.8) & 44.3 (±14.0) & \textbf{50.2} (±31.7) & 57.8 (±37.8) & 24.1 (±26.9) \\
                        & \modelpos-5   & 36.0 (±21.2) & 61.8 (±23.9) & \textbf{51.1} (±17.4) & 43.7 (±07.4) & \textbf{62.6} (±21.6) & \textbf{43.9} (±38.3) \\\bottomrule
\end{tabular}
\caption{Results of \model averaged over four templates using RoBERTa-large as backbone.}\label{tab:atop_roberta}
\end{table*}

\begin{table*}[tb]
\centering
\small
\begin{tabular}{lrrrrrrr}
\toprule
Metric                  & Trigger       & FR           & FN           & HATE         & IMDB         & SST          & AG           \\\midrule
CACC                    & -             & 84.0 (±02.6) & 72.7 (±06.0) & 78.8 (±06.2) & 80.3 (±03.1) & 82.1 (±04.4) & 86.5 (±01.4) \\\midrule
ASR                     & \modelall-3   & 32.1 (±14.0) & 35.8 (±12.0) & 33.2 (±23.0) & 13.9 (±17.1) & 45.8 (±20.8) & 17.8 (±16.2) \\
                        & \modelpos-3   & 28.1 (±15.2) & 46.3 (±14.4) & \textbf{48.0} (±25.4) & \textbf{21.8} (±32.8) & \textbf{57.3} (±27.0) & 30.5 (±28.0) \\\cmidrule(lr){2-2}
                        & \modelall-5   & \textbf{38.3} (±27.2) & 38.1 (±10.0) & 36.6 (±18.6) & 14.2 (±19.9) & 47.6 (±24.6) & 24.9 (±16.9) \\
                        & \modelpos-5   & \textbf{38.3} (±16.0) & \textbf{47.7} (±14.0) & 47.6 (±29.0) & 18.6 (±28.2) & 49.4 (±21.5) & \textbf{45.9} (±28.7) \\\bottomrule
\end{tabular}
\caption{Transferability of \model. Here we attack PFTs backboned with the BERT-large using triggers on RoBERTa-large. Results are averaged over four templates.  }\label{tab:atop_bert}
\end{table*}

\begin{table*}[tb]
\centering
\small
\begin{tabular}{lrrrrrrr}
\toprule
Metric                  & Trigger       & FR           & FN           & HATE         & IMDB         & SST          & AG           \\\midrule
CAAC                    & -            & 85.5 (±03.9) & 86.2 (±03.7) & 81.5 (±05.1) & 80.0 (±04.5) & 78.1 (±00.3) & 86.1 (±00.2) \\\midrule
ASR                     & RAND-3       & 5.8 (±01.1)  & 1.6 (±00.6) & 4.5 (±01.5)  & 7.0 (±02.9)  & 7.7 (±01.7)  & 2.0 (±00.7) \\
                        & \modelft-3    & 3.8 (±00.7)  & 2.1 (±00.3)  & 4.2 (±00.9)  & 5.5 (±03.1)  & 6.3 (±00.8)  & 2.2 (±00.5)  \\\cmidrule(lr){2-2}
                        & RAND-5        & 11.0 (±02.7) & 2.6 (±01.7) & 6.4 (±02.3)  & 8.1 (±04.1)  & 10.8 (±03.6) & 3.0 (±01.8) \\
                        & \modelft-5    & \textbf{14.6} (±10.8) & \textbf{2.9} (±00.7)  & \textbf{10.0} (±06.0) & \textbf{10.5} (±05.1) & \textbf{12.0} (±05.7) & \textbf{5.8} (±03.7)  \\\bottomrule
\end{tabular}
\caption{Results of \modelft on FT with the RoBERTa-large as backbone.}\label{tab:atoft}
\end{table*}

\begin{figure}[tb]
    \centering
    \includegraphics[width=\columnwidth]{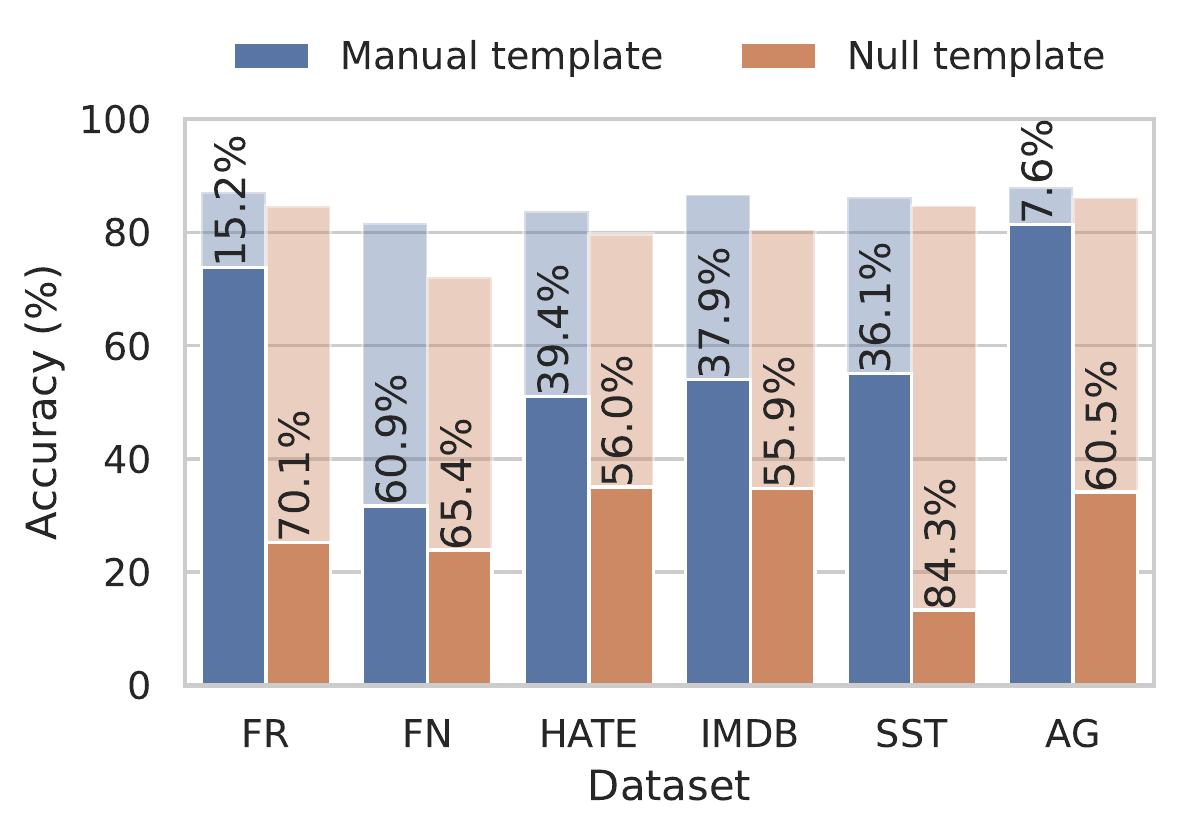}
    \caption{Comparing CACC and after-attack accuracy on different types of templates. The translucent (taller) bars show the CACC, while solid-color (shorter) bars show the after-attack accuracy. The value on each bar is ASR.}
    \label{fig:barchart_atop_prompt}
\end{figure}

\begin{figure}[tb]
    \centering
    \includegraphics[width=\columnwidth]{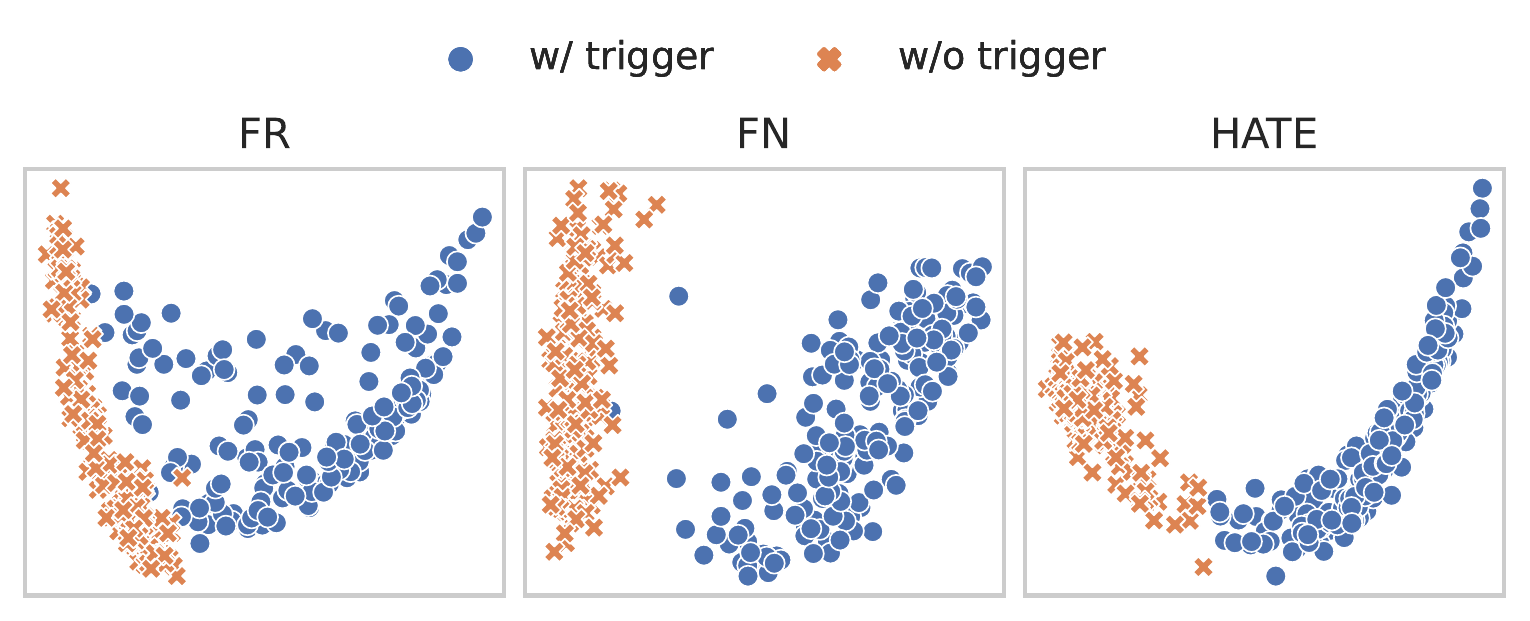}
    \includegraphics[width=\columnwidth]{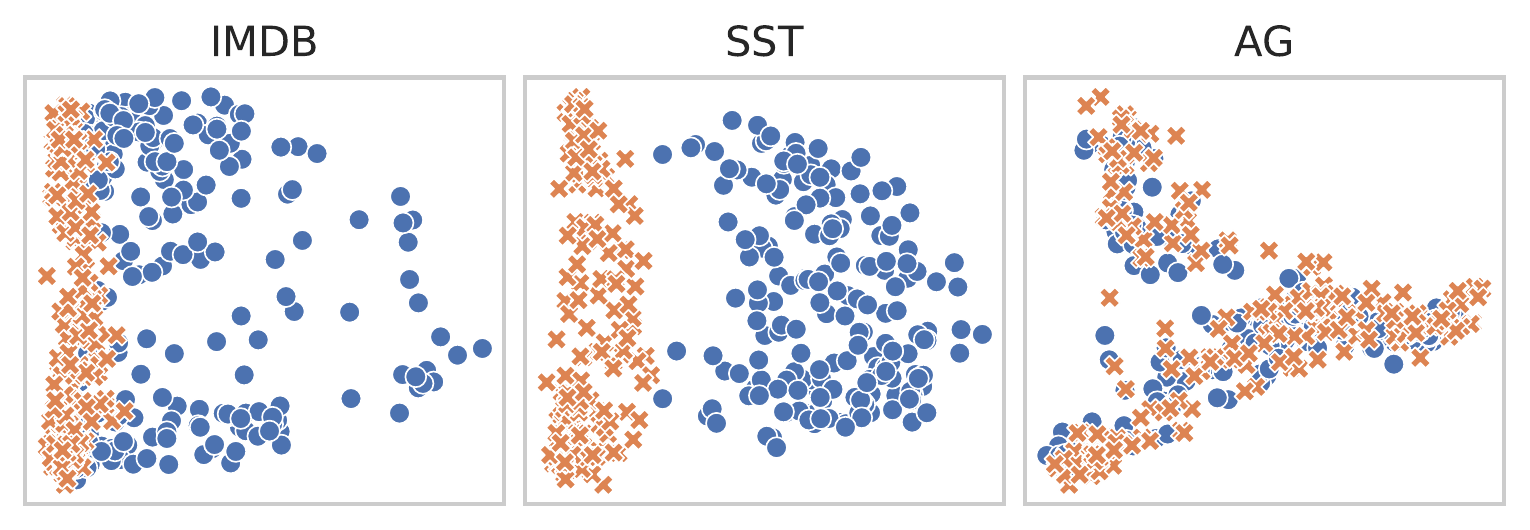}
    \caption{Visualization of the \textit{<mask>} embedding with and without trigger. Here we use ``Code Videos Replay <iframe'' from \modelall-5, and evaluate it on a manual template.}
    \label{fig:visulize_atop}
\end{figure}

\subsection{Triggers Discovered on RoBERTa}
The trigger sets we found are shown in Appendix. Some examples are ``Email Cancel Send'', and ``Code Videos Replay \textless{}iframe''. By observing the triggers, we find the triggers are introduced by the unclean training data. Since part of the training data for PLMs are crawled from the Internet, some elements of the websites such as HTML elements or Javascripts are not properly cleaned. Therefore, PLMs may learn spurious correlations. \model takes advantage of these elements to construct triggers.

\subsection{\model Attack Results}
Table~\ref{tab:atop_roberta} shows the performance of \model. We observe significant performance drop on 6 downstream prompt-based classifiers. The average attack success rate for \modelpos-5 is 49.9\%, significantly better than the random baseline. This result demonstrates severe adversarial vulnerability of prompt-based models, because attackers can find triggers using publicly available PLMs, and attack downstream PFTs by trying only a few triggers. As expected, 5-token triggers are more effective than 3-token triggers. We also find position sensitivity is more helpful for 3-token triggers.

We break down the results by the prompt type on Figure~\ref{fig:barchart_atop_prompt} and by relative position of \textit{<mask>} and \textit{<text>} in Appendix~\ref{appdix:exp2}. We found that manual templates are more robust than null templates, while the relative position of \textit{<mask>} and \textit{<text>} shows an ambiguous impact on ASRs.

We further investigate the behavior of prompt-based classifiers. We use PCA to reduce the dimension of the language model output on the \textit{<mask>} token and visualize it on Figure~\ref{fig:visulize_atop}. We found in most cases, the \textit{<mask>} embeddings are also shifted significantly after inserting the trigger. However the degree of the shift is less than backdoor triggers.

\begin{figure}[tb]
    \centering
    \includegraphics[width=\columnwidth]{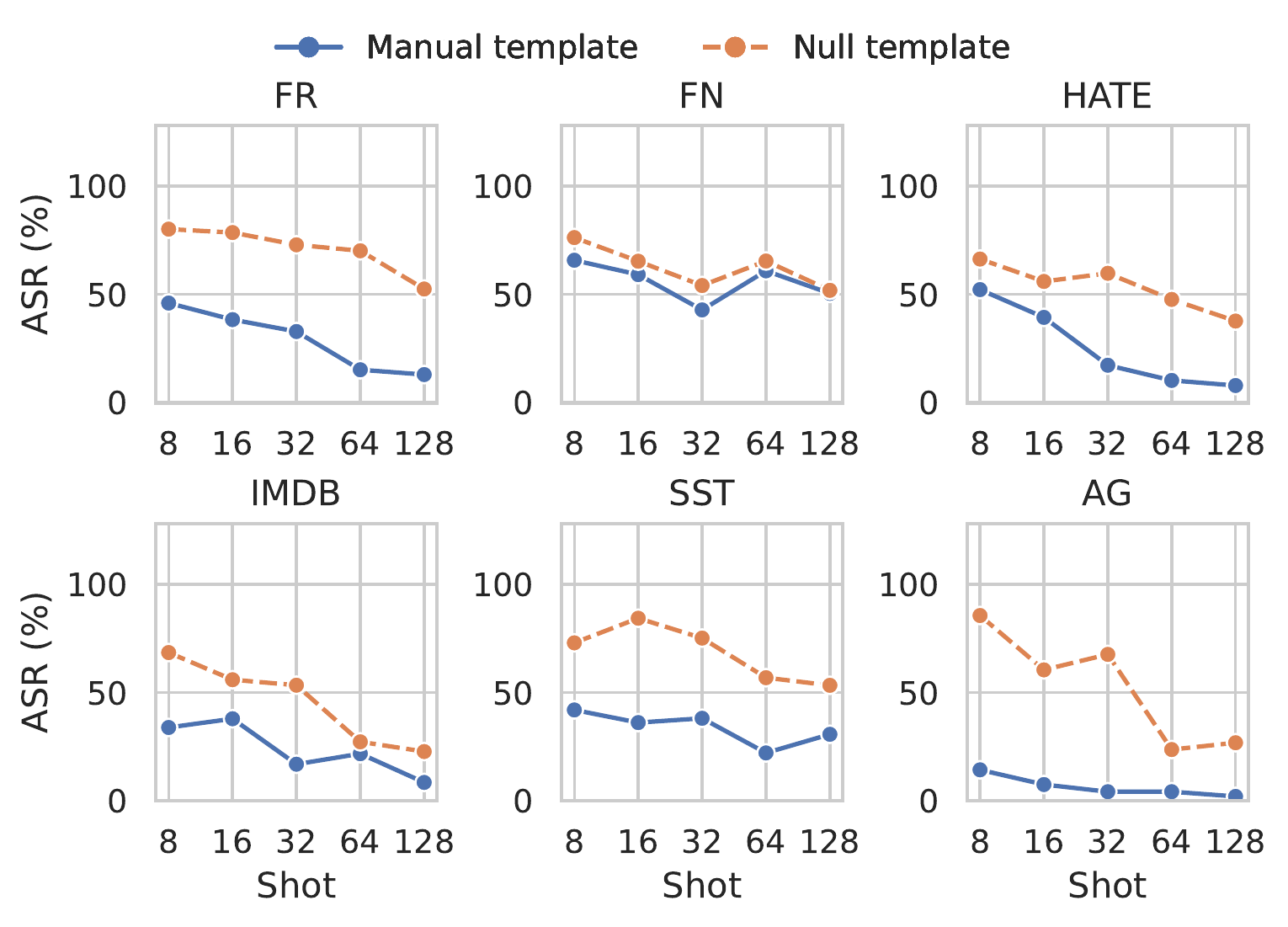}
    \caption{Comparing different shots.}
    \label{fig:atop_shot}
\end{figure}

Figure~\ref{fig:atop_shot} shows the ASR when PFTs are trained with more shots. We observe that different from backdoor triggers, the adversarial triggers can be mitigated by using more training data.

\subsection{Trigger Transferability}
\model is tied to a specific PLM. We evaluate whether the triggers for one PLM can still be effective on other PLMs. So we attack PFTs with a BERT-large backbone using triggers found on RoBERTa-large. The attack results on Table~\ref{tab:atop_bert} show that \model has strong transferability, and \modelpos is more effective after transferring to another PLM. But the advantage of longer triggers diminishes in transfer.

\begin{figure}[htb]
    \centering
    \includegraphics[width=\columnwidth]{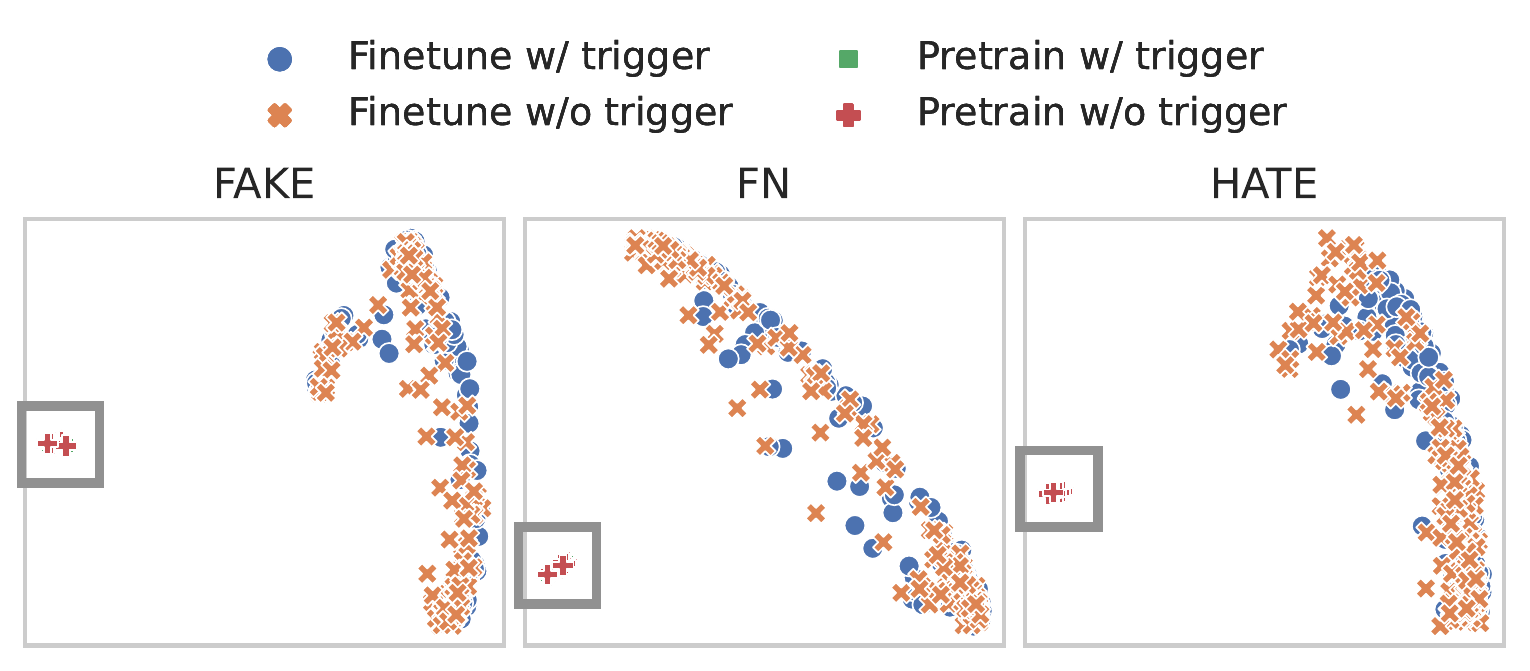}
    \includegraphics[width=\columnwidth]{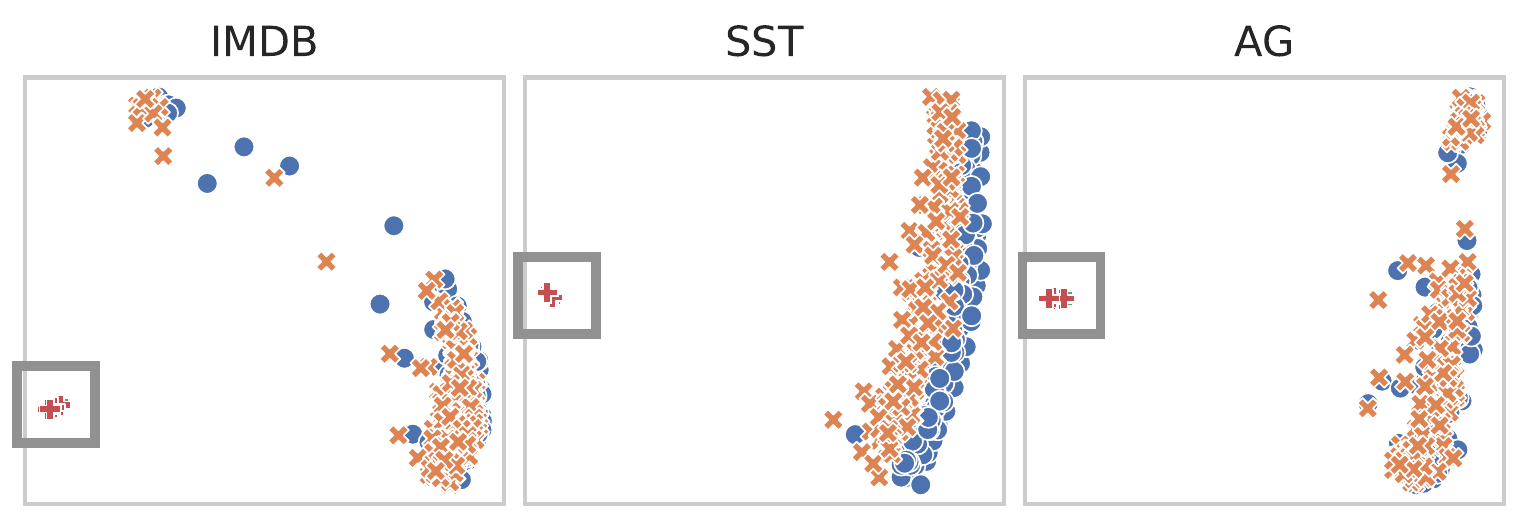}
    \caption{Visualization of the \textit{<cls>} embedding on FTs. Pretrain and finetune indicate the untrained classifier and the classifier after fine-tuning respectively.}
    \label{fig:atoft}
\end{figure}

\subsection{Compare with Fine-tuned Models}
We evaluate if FTs also suffer from adversarial triggers from PLMs. 
We adapt \model to FTs and named it \modelft. We search for \modelft such that it can best change the output embedding of the \textit{<cls>} token in the PLM. And we use the set of triggers to attack downstream FTs. (See Appendix~\ref{appdix:ft} for details.) Table~\ref{tab:atoft} shows that \modelft marginally outperforms random triggers. We also visualize the embeddings for the \textit{<cls>} token on Figure~\ref{fig:atoft}. We observe that injecting the trigger does not affect the \textit{<cls>} embedding much, while the embedding has a drastic shift before and after fine-tuning. It shows that traditional fine-tuning causes the shift of \textit{<cls>} embedding thus degenerates the efficacy of triggers. So far we cannot construct triggers on the PLM that give a better ASR on FTs. % believe the universal vulnerability is a unique phenomenon in prompt-based learning. 

\section{Mitigating the Universal Vulnerability}
Given the success of our attack methods, we propose a unified defense method based on outlier filtering against them.
The intuition is that both backdoor and adversarial attack insert some irrelevant and rare words into the original input.
Thus, a well-trained language model may detect these outlier words based on contextual information.
Our method is inspired by ONION~\cite{qi2020onion}, and simplifies it so that a held-out validation set is not required.
Given the input $\bx = [x_1, ..., x_i, ..., x_n]$, where $x_i$ is the $i$-th word in $\bx$. We propose to remove $x_i$ if removing it leads to a lower perplexity. We measure perplexity using GPT2-large. Table~\ref{tab:defense} shows the defense results.

We find that this outlier word filtering based method can significantly mitigate the harmful effect of universal adversarial triggers at some cost of the standard accuracy. 
However, the effect of defense against backdoor triggers is limited. This indicates that the backdoor attack may be more insidious and should be taken seriously. 

%               & HATE         & SST          \\\midrule
% CACC            & 76.8 (±03.6) & 83.0 (±02.6) \\
% \modelbd    & 87.9 (±10.5) & 79.7 (±19.9) \\
% \modelall-3    & 11.5 (±05.3) & 8.4 (±06.1)  \\
% \modelpos-3    & 17.2 (±09.6) & 18.8 (±12.1) \\
% \modelall-5    & 19.5 (±14.8) & 17.3 (±21.0) \\
% \modelpos-5    & 17.9 (±13.1) & 14.4 (±07.9) \\

\begin{table}[htb]
\centering\small
\begin{tabular}{rr@{\hskip 1ex}rr@{\hskip 1ex}r}
\toprule
           & \multicolumn{2}{c}{HATE (CACC -5.0\%)} & \multicolumn{2}{c}{SST (CACC -2.5\%)} \\\cmidrule(lr){2-3} \cmidrule(lr){4-5} 
Trigger    & ASR (\%)   & $\Delta$ (\%)  & ASR (\%)   & $\Delta$ (\%) \\\midrule
\modelbd    & 87.9 (±10.5)   & -11.7   & 79.7 (±19.9)   & -20.2  \\\midrule
\modelall-3 & 11.5 (±05.3)   & -24.0   & 8.4 (±06.1)    & -17.7  \\
\modelpos-3 & 17.2 (±09.6)   & -28.1   & 18.8 (±12.1)   & -14.6  \\
\modelall-5 & 19.5 (±14.8)   & -24.8   & 17.3 (±21.0)   & -40.5  \\
\modelpos-5 & 17.9 (±13.1)   & -33.2   & 14.4 (±07.9)   & -48.2  \\
\bottomrule
\end{tabular}
\caption{ASR after applying the outlier word filtering. $\Delta$ indicates the change of ASR.}
\label{tab:defense}
\end{table}

\section{Related Works}
\paragraph{Prompt-based Learning}
% The conventional fine-tuning method uses the special \textit{<cls>} token to the input text and deem the representation embedding of this token as the sentence representation.
% Then a classification head is employed to make the prediction, given the sentence representation as the input. The classification head and the language model are fine-tuned together on the down-stream tasks~\cite{Devlin2019BERTPO}.

Prompt-based learning paradigm in PLM fine-tuning has emerged recently and been intensively studied, especially in the few-shot setting~\citep{liu2021promptsurvey}. 
These methods reformulate the classification task as a blank-filling task by wrapping the original texts with templates that contain \textit{<mask>} tokens. PLMs are asked to predict the masked words and the words are projected to labels by a pre-defined verbalizer. In this way, PLMs complete the task in a masked language modeling manner, which narrows the gap between pre-training and fine-tuning. 
% using the mask token and the language model head. 
%
There are various sorts of prompts, including manually designed ones~\citep{Brown2020GPT3, petroni2019lama, schick2021exploiting}, automatically searched ones~\citep{shin2020autoprompt, gao2021making}, and continuously optimized ones~\citep{li2021prefixtuning, lester2021prompttuning}. Among them, manual prompts share the highest similarity with pre-training, because they adopt human-understandable templates to stimulate PLMs.
%Continuous prompt methods are closely related to parameter efficient tuning \citep{houlsby2019adapter, li2021prefixtuning, lester2021prompttuning}. 
%\citet{liu2021promptsurvey} gives a comprehensive survey on prompt-based methods.  
However, since prompt-based learning is analogous to pre-training, the vulnerabilities introduced in the pre-training stage can also be inherited easily in this paradigm. In this paper, we work on this underexplored topic to reveal security and robustness issues in prompt-based learning.

\paragraph{Backdoor Attack}
The backdoor attack is less investigated in NLP. Recent work usually implants backdoors through data poisoning. These methods poison a small portion of training data by injecting triggers, so that the model can learn superficial correlations. According to the form of the trigger, it can be categorized as poisoning in the input space where irrelevant words or sentences are injected into the original text~\cite{kurita-etal-2020-weight, dai2019backdoor, chen2020badnl}; and poisoning in feature space where the syntax pattern or the style of the text is modified~\cite{qi-etal-2021-hidden, qi2021mind}. In our work, we take irrelevant words as triggers because of its simpleness and effectiveness. 

%Current works mostly inject backdoors by poisoning a small portion of  training dataset. Depending on the format of the backdoor triggers, they can be categorized as visible triggers and invisible triggers. Visible triggers can be pre-defined irrelevant words (e.g. "bb", "mn") or sentences (e.g. "I love watching this movie") \citep{kurita-etal-2020-weight, dai2019backdoor, chen2020badnl}.

% Current works mostly explore the training dataset poisoning and can be roughly classified into two categories, depending on the attack space. Specifically, this is related to the employed backdoor triggers.
%

%The first kind of triggers are directly injected into the surface space. Attackers may inject irrelevant words ("bb", "mn") or sentence ("I love watching this movie") into the original sentence to craft poison samples~\citep{kurita-etal-2020-weight, dai2019backdoor, chen2020badnl}.
%
%Besides, there are also works exploring invisible triggers injected into the feature space, including specific syntax patterns \citep{qi-etal-2021-hidden} and text styles \citep{qi2021mind}. We employ the first kind of trigger in this work because of its effectiveness.

\paragraph{Adversarial Attack} 
Adversarial vulnerability is a known issue for deep-learning-based models.
There are a number of attack methods being proposed, including character-level methods \citep{li2019textbugger}, word-level methods \citep{ren2019pwws, jin2019textfooler, zang2019sememepso}, sentence-level methods \cite{qi2021mind, wang2020cat, xu2021attacking}, and multi-granularity methods \cite{wang2019t3, chen2021multi}. 
% Recent attack methods use PLMs to propose word perturbations \citep{garg2020bae, li2021clare}, so that the adversarial sentences can be more fluent. 
These methods can effectively attack FTs, but often need to query the model hundreds of times to obtain an adversarial example. Universal adversarial trigger~\citep{wallace2019uat} is an attempt to reduce the number of queries and construct a more general trigger that is effective on multiple examples. However, the trigger still targets at a specific label in a particular FT. We emphasize that this approach differs from \model in that our method focuses on the new prompt-based learning paradigm, and our triggers are applicable to arbitrary labels in arbitrary PFTs, thus being more universal.
%

%Our work is motivated by universal adversarial trigger \cite{wallace2019uat}, a method that searches short phrases to interrupt the prediction of fine-tuned classifiers. 

\section{Conclusion}
We explore the universal vulnerabilities of prompt-based learning paradigm from the backdoor attack and the adversarial attack perspectives, depending on whether the attackers can control the pre-training stage.
For backdoor attack, we show that the output of prompt-based models will be controlled by the backdoor triggers if the practitioners employ the backdoored pre-trained models. 
For adversarial attack, we show that the performance of prompt-based models decreases if the input text is inserted into adversarial triggers, which are constructed from only the plain texts.
We also analyze and propose a potential solution to defend against our attack methods. 
Through this work, we call on the research community to pay more attention to the universal vulnerabilities of the prompt-based learning paradigm before it is widely adopted.

% We show that prompt-based learning preserves the adversarial triggers in the pre-trained language model. It is a severe security issue as malicious attackers can derive adversarial examples easily on pre-trained language models and conduct attack on all downstream prompt-based models. The traditional fine-tune-based classifiers do not have similar problems. Therefore, we call on the research community to pay more attention to the unique vulnerabilities of the prompt-based learning. 

\section*{Ethical Consideration}
In this paper, we take the position of an attacker, and propose to conduct a backdoor attack and adversarial attack against PFTs. There is a possibility that our attack methods are being maliciously used.
However, research on attacks against PFTs is still necessary and very important for two reasons: (1) we can gain insights from the experimental results, that can help us defend against the proposed attacks, and design better prompt-based models; (2) we reveal the universal vulnerability of the prompt-based learning paradigm, so that practitioners understand the potential risk when deploying these models.

% In this section, we discuss the ethical considerations of our paper.

% \paragraph{Intended Use} In this paper, we propose to conduct backdoor attack and adversarial attack against PFTs from the standpoint of an attacker. Our motivations are: (1) we can gain insights from the experimental results, that can help us defend against the proposed attacks, and design better prompt-based models; (2) we show the universal vulnerability of the prompt-based learning paradigm, so that people understand the potential risk of deploying these models. The proposed attack methods should be used to improve prompt-based learning paradigm rather than attacking real systems.

% \paragraph{Potential Risk.} There is a possibility that our attack methods being maliciously abused.
% However, research on backdoor attack and adversarial attack against PFTs is still necessary and very important since it helps the research community understand powerful attack methods and the potential risks so that defending methods can be better designed.

% \paragraph{Energy saving. } We list the settings of hyper-parameters of our method in Appendix, to prevent people from conducting unnecessary tuning and help researchers to quickly reproduce our results. We will also release the checkpoints including all victim models to avoid repeated energy costs.

% Entries for the entire Anthology, followed by custom entries
\bibliography{ref}

\begin{thebibliography}{35}
\expandafter\ifx\csname natexlab\endcsname\relax\def\natexlab#1{#1}\fi

\bibitem[{Brown et~al.(2020)Brown, Mann, Ryder, Subbiah, Kaplan, Dhariwal,
  Neelakantan, Shyam, Sastry, Askell, Agarwal, Herbert-Voss, Krueger, Henighan,
  Child, Ramesh, Ziegler, Wu, Winter, Hesse, Chen, Sigler, Litwin, Gray, Chess,
  Clark, Berner, McCandlish, Radford, Sutskever, and Amodei}]{Brown2020GPT3}
Tom Brown, Benjamin Mann, Nick Ryder, Melanie Subbiah, Jared~D Kaplan, Prafulla
  Dhariwal, Arvind Neelakantan, Pranav Shyam, Girish Sastry, Amanda Askell,
  Sandhini Agarwal, Ariel Herbert-Voss, Gretchen Krueger, Tom Henighan, Rewon
  Child, Aditya Ramesh, Daniel Ziegler, Jeffrey Wu, Clemens Winter, Chris
  Hesse, Mark Chen, Eric Sigler, Mateusz Litwin, Scott Gray, Benjamin Chess,
  Jack Clark, Christopher Berner, Sam McCandlish, Alec Radford, Ilya Sutskever,
  and Dario Amodei. 2020.
\newblock Language models are few-shot learners.
\newblock In \emph{NeurIPS}.

\bibitem[{Chen et~al.(2021{\natexlab{a}})Chen, Salem, Backes, Ma, and
  Zhang}]{chen2020badnl}
Xiaoyi Chen, Ahmed Salem, Michael Backes, Shiqing Ma, and Yang Zhang.
  2021{\natexlab{a}}.
\newblock Badnl: Backdoor attacks against nlp models.
\newblock In \emph{ICML Workshop}.

\bibitem[{Chen et~al.(2021{\natexlab{b}})Chen, Su, and Wei}]{chen2021multi}
Yangyi Chen, Jin Su, and Wei Wei. 2021{\natexlab{b}}.
\newblock Multi-granularity textual adversarial attack with behavior cloning.
\newblock \emph{arXiv preprint}.

\bibitem[{Dai et~al.(2019)Dai, Chen, and Li}]{dai2019backdoor}
Jiazhu Dai, Chuanshuai Chen, and Yufeng Li. 2019.
\newblock A backdoor attack against lstm-based text classification systems.
\newblock \emph{IEEE Access}.

\bibitem[{Devlin et~al.(2019)Devlin, Chang, Lee, and
  Toutanova}]{Devlin2019BERTPO}
Jacob Devlin, Ming-Wei Chang, Kenton Lee, and Kristina Toutanova. 2019.
\newblock Bert: Pre-training of deep bidirectional transformers for language
  understanding.
\newblock In \emph{NAACL}.

\bibitem[{Gao et~al.(2021)Gao, Fisch, and Chen}]{gao2021making}
Tianyu Gao, Adam Fisch, and Danqi Chen. 2021.
\newblock Making pre-trained language models better few-shot learners.
\newblock In \emph{ACL}.

\bibitem[{Gulli()}]{gulliAGdataset}
Antonio Gulli.
\newblock Ag's corpus of news articles.

\bibitem[{Jin et~al.(2020)Jin, Jin, Zhou, and Szolovits}]{jin2019textfooler}
Di~Jin, Zhijing Jin, Joey~Tianyi Zhou, and Peter Szolovits. 2020.
\newblock Is bert really robust? natural language attack on text classification
  and entailment.
\newblock In \emph{AAAI}.

\bibitem[{Kurita et~al.(2020{\natexlab{a}})Kurita, Michel, and
  Neubig}]{kurita20acl}
Keita Kurita, Paul Michel, and Graham Neubig. 2020{\natexlab{a}}.
\newblock Weight poisoning attacks on pretrained models.
\newblock In \emph{ACL}.

\bibitem[{Kurita et~al.(2020{\natexlab{b}})Kurita, Michel, and
  Neubig}]{kurita-etal-2020-weight}
Keita Kurita, Paul Michel, and Graham Neubig. 2020{\natexlab{b}}.
\newblock Weight poisoning attacks on pretrained models.
\newblock In \emph{ACL}.

\bibitem[{Lester et~al.(2021)Lester, Al-Rfou, and
  Constant}]{lester2021prompttuning}
Brian Lester, Rami Al-Rfou, and Noah Constant. 2021.
\newblock The power of scale for parameter-efficient prompt tuning.
\newblock In \emph{EMNLP}.

\bibitem[{Li et~al.(2019)Li, Ji, Du, Li, and Wang}]{li2019textbugger}
J~Li, S~Ji, T~Du, B~Li, and T~Wang. 2019.
\newblock Textbugger: Generating adversarial text against real-world
  applications.
\newblock In \emph{Annual Network and Distributed System Security Symposium}.

\bibitem[{Li et~al.(2021)Li, Liu, Dong, Zhao, Xue, Zhu, and
  Lu}]{Li2021HiddenBI}
Shaofeng Li, Hui Liu, Tian Dong, Benjamin Zi~Hao Zhao, Minhui Xue, Haojin Zhu,
  and Jialiang Lu. 2021.
\newblock Hidden backdoors in human-centric language models.
\newblock In \emph{ACM SIGSAC Conference on Computer and Communications
  Security}.

\bibitem[{Li and Liang(2021)}]{li2021prefixtuning}
Xiang~Lisa Li and Percy Liang. 2021.
\newblock Prefix-tuning: Optimizing continuous prompts for generation.
\newblock In \emph{ACL-IJCNLP}.

\bibitem[{Liu et~al.(2021)Liu, Yuan, Fu, Jiang, Hayashi, and
  Neubig}]{liu2021promptsurvey}
Pengfei Liu, Weizhe Yuan, Jinlan Fu, Zhengbao Jiang, Hiroaki Hayashi, and
  Graham Neubig. 2021.
\newblock Pre-train, prompt, and predict: A systematic survey of prompting
  methods in natural language processing.
\newblock \emph{arXiv preprint}.

\bibitem[{Logan~IV et~al.(2021)Logan~IV, Bala{\v{z}}evi{\'c}, Wallace, Petroni,
  Singh, and Riedel}]{logan2021cutting}
Robert~L Logan~IV, Ivana Bala{\v{z}}evi{\'c}, Eric Wallace, Fabio Petroni,
  Sameer Singh, and Sebastian Riedel. 2021.
\newblock Cutting down on prompts and parameters: Simple few-shot learning with
  language models.
\newblock \emph{arXiv preprint}.

\bibitem[{Loshchilov and Hutter(2019)}]{loshchilov2018fixing}
Ilya Loshchilov and Frank Hutter. 2019.
\newblock Decoupled weight decay regularization.
\newblock In \emph{ICLR}.

\bibitem[{Maas et~al.(2011)Maas, Daly, Pham, Huang, Ng, and
  Potts}]{maas2011imdb}
Andrew~L Maas, Raymond~E Daly, Peter~T Pham, Dan Huang, Andrew~Y Ng, and
  Christopher Potts. 2011.
\newblock Learning word vectors for sentiment analysis.
\newblock In \emph{ACL-HLT}.

\bibitem[{Merity et~al.(2017)Merity, Xiong, Bradbury, and
  Socher}]{Merity2017wikitext}
Stephen Merity, Caiming Xiong, James Bradbury, and Richard Socher. 2017.
\newblock Pointer sentinel mixture models.
\newblock In \emph{ICLR}.

\bibitem[{Petroni et~al.(2019)Petroni, Rockt{\"a}schel, Riedel, Lewis, Bakhtin,
  Wu, and Miller}]{petroni2019lama}
Fabio Petroni, Tim Rockt{\"a}schel, Sebastian Riedel, Patrick Lewis, Anton
  Bakhtin, Yuxiang Wu, and Alexander Miller. 2019.
\newblock Language models as knowledge bases?
\newblock In \emph{EMNLP-IJCNLP}.

\bibitem[{Qi et~al.(2021{\natexlab{a}})Qi, Chen, Li, Yao, Liu, and
  Sun}]{qi2020onion}
Fanchao Qi, Yangyi Chen, Mukai Li, Yuan Yao, Zhiyuan Liu, and Maosong Sun.
  2021{\natexlab{a}}.
\newblock Onion: A simple and effective defense against textual backdoor
  attacks.
\newblock In \emph{EMNLP}.

\bibitem[{Qi et~al.(2021{\natexlab{b}})Qi, Chen, Zhang, Li, Liu, and
  Sun}]{qi2021mind}
Fanchao Qi, Yangyi Chen, Xurui Zhang, Mukai Li, Zhiyuan Liu, and Maosong Sun.
  2021{\natexlab{b}}.
\newblock Mind the style of text! adversarial and backdoor attacks based on
  text style transfer.
\newblock In \emph{EMNLP}.

\bibitem[{Qi et~al.(2021{\natexlab{c}})Qi, Li, Chen, Zhang, Liu, Wang, and
  Sun}]{qi-etal-2021-hidden}
Fanchao Qi, Mukai Li, Yangyi Chen, Zhengyan Zhang, Zhiyuan Liu, Yasheng Wang,
  and Maosong Sun. 2021{\natexlab{c}}.
\newblock Hidden killer: Invisible textual backdoor attacks with syntactic
  trigger.
\newblock In \emph{ACL-IJCNLP}.

\bibitem[{Ren et~al.(2019)Ren, Deng, He, and Che}]{ren2019pwws}
Shuhuai Ren, Yihe Deng, Kun He, and Wanxiang Che. 2019.
\newblock Generating natural language adversarial examples through probability
  weighted word saliency.
\newblock In \emph{ACL}.

\bibitem[{Salminen et~al.(2022)Salminen, Kandpal, Kamel, gyo Jung, and
  Jansen}]{SALMINEN2022102771}
Joni Salminen, Chandrashekhar Kandpal, Ahmed~Mohamed Kamel, Soon gyo Jung, and
  Bernard~J. Jansen. 2022.
\newblock Creating and detecting fake reviews of online products.
\newblock \emph{Journal of Retailing and Consumer Services}.

\bibitem[{Schick and Sch{\"u}tze(2021)}]{schick2021exploiting}
Timo Schick and Hinrich Sch{\"u}tze. 2021.
\newblock Exploiting cloze-questions for few-shot text classification and
  natural language inference.
\newblock In \emph{EACL}.

\bibitem[{Shin et~al.(2020)Shin, Razeghi, Logan~IV, Wallace, and
  Singh}]{shin2020autoprompt}
Taylor Shin, Yasaman Razeghi, Robert~L Logan~IV, Eric Wallace, and Sameer
  Singh. 2020.
\newblock Eliciting knowledge from language models using automatically
  generated prompts.
\newblock In \emph{EMNLP}.

\bibitem[{Wallace et~al.(2019)Wallace, Feng, Kandpal, Gardner, and
  Singh}]{wallace2019uat}
Eric Wallace, Shi Feng, Nikhil Kandpal, Matt Gardner, and Sameer Singh. 2019.
\newblock Universal adversarial triggers for attacking and analyzing nlp.
\newblock In \emph{EMNLP-IJCNLP}.

\bibitem[{Wang et~al.(2019{\natexlab{a}})Wang, Singh, Michael, Hill, Levy, and
  Bowman}]{wang2018glue}
Alex Wang, Amanpreet Singh, Julian Michael, Felix Hill, Omer Levy, and Samuel~R
  Bowman. 2019{\natexlab{a}}.
\newblock Glue: A multi-task benchmark and analysis platform for natural
  language understanding.
\newblock In \emph{ICLR}.

\bibitem[{Wang et~al.(2019{\natexlab{b}})Wang, Pei, Pan, Chen, Wang, and
  Li}]{wang2019t3}
Boxin Wang, Hengzhi Pei, Boyuan Pan, Qian Chen, Shuohang Wang, and Bo~Li.
  2019{\natexlab{b}}.
\newblock T3: Tree-autoencoder constrained adversarial text generation for
  targeted attack.
\newblock \emph{arXiv preprint}.

\bibitem[{Wang et~al.(2020)Wang, Wang, Qin, Packer, Li, Chen, Beutel, and
  Chi}]{wang2020cat}
Tianlu Wang, Xuezhi Wang, Yao Qin, Ben Packer, Kang Li, Jilin Chen, Alex
  Beutel, and Ed~Chi. 2020.
\newblock Cat-gen: Improving robustness in nlp models via controlled
  adversarial text generation.
\newblock \emph{arXiv preprint}.

\bibitem[{Xu and Veeramachaneni(2021)}]{xu2021attacking}
Lei Xu and Kalyan Veeramachaneni. 2021.
\newblock Attacking text classifiers via sentence rewriting sampler.
\newblock \emph{arXiv preprint}.

\bibitem[{Yang et~al.(2017)Yang, Mukherjee, and
  Dragut}]{yang-etal-2017-satirical}
Fan Yang, Arjun Mukherjee, and Eduard Dragut. 2017.
\newblock Satirical news detection and analysis using attention mechanism and
  linguistic features.
\newblock In \emph{EMNLP}.

\bibitem[{Zang et~al.(2020)Zang, Qi, Yang, Liu, Zhang, Liu, and
  Sun}]{zang2019sememepso}
Yuan Zang, Fanchao Qi, Chenghao Yang, Zhiyuan Liu, Meng Zhang, Qun Liu, and
  Maosong Sun. 2020.
\newblock Word-level textual adversarial attacking as combinatorial
  optimization.
\newblock In \emph{ACL}.

\bibitem[{Zhang et~al.(2021)Zhang, Xiao, Li, Lv, Qi, Liu, Wang, Jiang, and
  Sun}]{NeuBA}
Zhengyan Zhang, Guangxuan Xiao, Yongwei Li, Tian Lv, Fanchao Qi, Zhiyuan Liu,
  Yasheng Wang, Xin Jiang, and Maosong Sun. 2021.
\newblock Red alarm for pre-trained models: Universal vulnerability to
  neuron-level backdoor attacks.
\newblock \emph{arXiv preprint}.

\end{thebibliography}
\bibliographystyle{acl_natbib}

\clearpage
\appendix
\section{Pre-defined Embeddings for Backdoor Attack} \label{appdix:a}
In RoBERTa-large, the output is a 1024-dimensional embedding. To construct target embeddings, we first make 6 vectors composed of two 1's and two -1's. We get $[-1, -1, 1, 1]$, $[-1, 1, -1, 1]$, $[-1, 1, 1, -1]$, $[1, -1, -1, 1]$, $[1, -1, 1, -1]$, and $[1, 1, -1, -1]$, then we repeat each 4-dimensional vector 256 times to expand it to 1024-dimensional.

% We use the following code to generate the pre-defined embeddings.

% \begin{minted}[mathescape, breaklines, fontsize=\footnotesize, python3, frame=single]{python}
% trigger_li = ["cf", "mn", "bb", "qt", "pt", 'mt']
% hidden_size = 1024
% poison_labels = [[1] * hidden_size for _ in range(len(trigger_li))]
% i = 0
% for j in range(4):
%     for k in range(j + 1, 4):
%         for m in range(0, hidden_size // 4):
%             poison_labels[i][j * hidden_size // 4 + m] = -1
%             poison_labels[i][k * hidden_size // 4 + m] = -1
%         i += 1

% \end{minted}

\section{Adversarial Attack on FTs}\label{appdix:ft}
We adapt the idea of \model onto FTs and named it \modelft. Specifically, we modifies Eq.~\ref{eq:obj}, and tries two objectives.

(1) We first try to find a trigger that minimize the likelihood of the PLM to predict the \textit{<cls>} token in the input as itself, i.e. 
    \begin{equation}
    \text{minimize } \sum_{\bx\in\mathcal{D}} \log \mathcal{F_O}(\bx, \bt)_\text{<cls>},\label{eq:fta}
    \end{equation}
    where $\mathcal{F_O}(\bx, \bt)_\text{<cls>}$ is the probability of \textit{<cls>} being predicted as \textit{<cls>}.

(2) According to our observation on Figure~\ref{fig:visulize_atop}, we directly maximize the embedding shift on the \textit{<cls>} token when inserting the trigger, specifically
    \begin{equation}
    \text{maximize } \sum_{\bx\in\mathcal{D}} ||\mathcal{F_O}(\bx, \phi) - \mathcal{F_O}(\bx, \bt)||_2, \label{eq:ftb}
    \end{equation}
    where $\mathcal{F_O}(\bx, \bt)$ is the embedding of the \textit{<cls>} token when $\bt$ is injected, and $\phi$ means not using a trigger.
We report the result of Eq.~\ref{eq:ftb} in Table~\ref{tab:atoft}. 

\section{Additional Experimental Results}\label{appdix:exp}

\subsection{Results on backdoor attack}\label{appdix:exp1}
\begin{figure}[htb]
    \centering
    \includegraphics[width=\columnwidth]{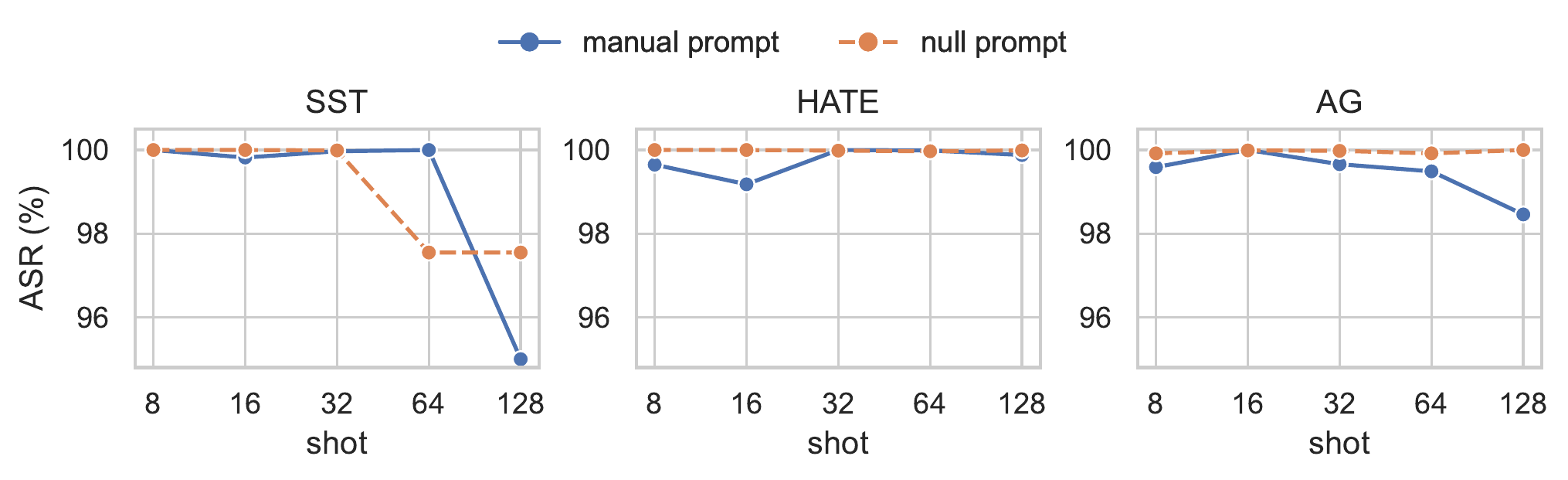}
    \caption{Comparing different shots.}
    \label{fig:atop_shot_backdoor}
\end{figure}

We experiment with different shots in backdoor attack. The results are listed in Figure~\ref{fig:atop_shot_backdoor}.

\subsection{Results on adversarial attack}\label{appdix:exp2}
\begin{figure}[htb]
    \centering
    \includegraphics[width=\columnwidth]{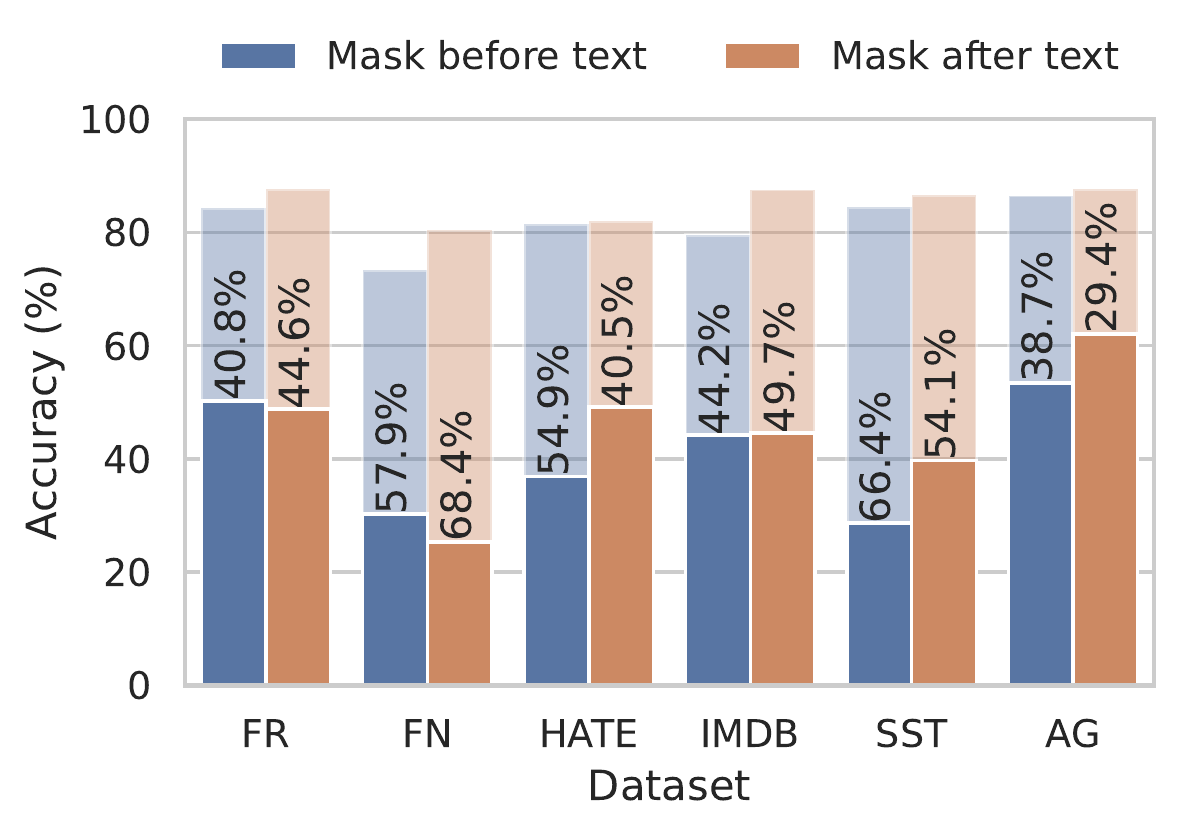}
    \caption{Comparing CACC and Attack Accuracy on the relative postion of \textit{<mask>} and \textit{<text>}. The translucent (taller) bars shows the CACC, while solid-color (shorter) bar shows the attack accuracy. The value on each bar is ASR.}
    \label{fig:barchart_atop_position}
\end{figure}

The triggers we found on RoBERTa-large is shown on Table~\ref{tab:triggers}. 
Figure~\ref{fig:barchart_atop_position} shows the effect of relative position of \textit{<mask>} and \textit{<text>} on ASR.
\begin{table}[htb]
\centering
\small
\begin{tabular}{ll}
\toprule
\textbf{Trigger set} & \textbf{Triggers} \\\midrule
\modelall-3     & Videos Loading Replay                  \\
                & Details DMCA Share                     \\
                & Email Cancel Send                      \\\midrule
\modelpos-3  & Reading Below Alicia                   \\
MBT                & Copy Transcript Share                  \\
                & edit {]} As                            \\
\modelpos-3  & organisers Crimes Against              \\
MAT                & \textbackslash{}"The Last              \\
                & disorder.{[} edit                      \\\midrule
\modelall-5     & Code Videos Replay \textless{}iframe   \\
                & 249 autoplay CopyContent        \\
                & Photo Skipatos Caption Skip          \\\midrule
\modelpos-5 & Code Copy Replay WATCHED Share         \\
MBT                & Address Email Invalid OTHERToday        \\
                & Duty Online Reset Trailer Details      \\
\modelpos-5  & yourselvesShareSkip Disable JavaScript \\
MAT                & Davis-{[}\{Contentibility                    \\
                & {[}…{]} announSHIPEmail Address        \\\bottomrule
\end{tabular}
\caption{Triggers we found in each setup.}\label{tab:triggers}
\end{table}


\section{Prompt templates}\label{appdix:templates}
Table~\ref{tab:prompt} shows all the prompt templates and verbalizers.
% Please add the following required packages to your document preamble:
% \usepackage{multirow}
\begin{table*}[htb]
\centering\small
\begin{tabular}{rlll}
\toprule
\textbf{Dataset}      & \textbf{Type} & \textbf{Prompt}            & \textbf{Verbalizer}                                  \\\midrule
\multirow{4}{*}{FR}   & Null          & \emph{<mask>} \emph{<trigger>} \emph{<text>}              & \multirow{4}{*}{real/fake}                           \\
                      & Template      & \emph{<text>} \emph{<trigger>} \emph{<mask>}              &                                                      \\
                      & Manual        & [ \emph{<mask>} review ] \emph{<trigger>} \emph{<text>} &                                                      \\
                      & Template      & \emph{<text>} \emph{<trigger>} [ \emph{<mask>} review ] &                                                      \\\midrule
\multirow{4}{*}{RN}   & Null          & \emph{<mask>} \emph{<trigger>} \emph{<text>}              & \multirow{4}{*}{real/fake}                           \\
                      & Template      & \emph{<text>} \emph{<trigger>} \emph{<mask>}              &                                                      \\
                      & Manual        & It was \emph{<mask>} . \emph{<trigger>} \emph{<text>}   &                                                      \\
                      & Template      & \emph{<text>} \emph{<trigger>} It was \emph{<mask>} .   &                                                      \\\midrule
\multirow{4}{*}{HATE} & Null          & \emph{<mask>} \emph{<trigger>} \emph{<text>}              & \multirow{4}{*}{harmless/hate}                       \\
                      & Template      & \emph{<text>} \emph{<trigger>} \emph{<mask>}              &                                                      \\
                      & Manual        & [ \emph{<mask>} speech ] \emph{<trigger>} \emph{<text>} &                                                      \\
                      & Template      & \emph{<text>} \emph{<trigger>} [ \emph{<mask>} speech ] &                                                      \\\midrule
\multirow{4}{*}{IMDB} & Null          & \emph{<mask>} \emph{<trigger>} \emph{<text>}              & \multirow{4}{*}{bad/good}                            \\
                      & Template      & \emph{<text>} \emph{<trigger>} \emph{<mask>}              &                                                      \\
                      & Manual        & It was \emph{<mask>} . \emph{<trigger>} \emph{<text>}   &                                                      \\
                      & Template      & \emph{<text>} \emph{<trigger>} It was \emph{<mask>} .   &                                                      \\\midrule
\multirow{4}{*}{SST}  & Null          & \emph{<mask>} \emph{<trigger>} \emph{<text>}              & \multirow{4}{*}{bad/good}                            \\
                      & Template      & \emph{<text>} \emph{<trigger>} \emph{<mask>}              &                                                      \\
                      & Manual        & It was \emph{<mask>} . \emph{<trigger>} \emph{<text>}   &                                                      \\
                      & Template      & \emph{<text>} \emph{<trigger>} It was \emph{<mask>} .   &                                                      \\\midrule
\multirow{4}{*}{AG}   & Null          & \emph{<mask>} \emph{<trigger>} \emph{<text>}              & \multirow{4}{*}{politics/sports/business/technology} \\
                      & Template      & \emph{<text>} \emph{<trigger>} \emph{<mask>}              &                                                      \\
                      & Manual        & [ \emph{<mask>} news ] \emph{<trigger>} \emph{<text>}     &                                                      \\
                      & Template      & \emph{<text>} \emph{<trigger>} [ \emph{<mask>} news ]     &                                                      \\\bottomrule
\end{tabular}
\caption{Prompts and verbalizers. For each template, we also mark the position where the triggers are injected.}\label{tab:prompt}
\end{table*}

\section{Beam Search Algorithm for Adversarial Attack}\label{appdix:alg}
Algorithm~\ref{alg:suat} shows the beam search algorithm.
\begin{algorithm*}[htb]
\SetAlgoLined
\KwIn{
Processed text corpora $\mathcal{D}'$; trigger length $l$,
number of search steps $n$; batch size $m$; beam size $b$.}
\KwOut{$b$ triggers of length $l$.}
\BlankLine
$\mathtt{current\_beam}=[\mathtt{random\_init\_a\_trigger()}]$\;
\For{$i \in 1\ldots n$}{
    $\mathtt{new\_beam} = \text{empty list}$\;
    $[(\bx^{(j)}, y^{(j)})]_{j=1\ldots m} \sim \mathcal{D}'$\;
    \For{$k \in 1\ldots l$} {
        \For{$\bt \in \mathtt{current\_trigger}$} {
            $\mathtt{loss} = \sum_{j=1}^{m}\mathtt{compute\_loss}(\bx^{(j)}, y^{(j)}, \bt)$\;
            $\mathtt{new\_beam.add}((\bt, \mathtt{loss}))$\;
            $\mathtt{grad}=\nabla_{\mathtt{word\_embedding}(\bt_k)} \mathtt{loss}$\;
            $\mathtt{weight}_c=-\langle \mathtt{grad}, \mathtt{word\_embedding}(c) - \mathtt{word\_embedding}(t_i) \rangle$\;
            $\mathtt{candidate\_words} = \text{get }b\text{ words with maximum }\mathtt{weight}$\;
            \For {$c \in \mathtt{candidate\_words}$} {
                $\bt'=\bt_{1:k-1},c,\bt_{k+1:l}$\;
                $\mathtt{loss} = \sum_{u=1}^{m}\mathtt{compute\_loss}(\bx^{(u)}, y^{(u)}, \bt')$\; 
                $\mathtt{new\_beam.add}((\bt', \mathtt{loss}))$\;
            }
        }
        $\mathtt{current\_beam}= \text{get }b\text{ best triggers from }\mathtt{new\_beam}$\;
    }
}
\Return{$\mathtt{current\_beam}$}
\caption{Beam Search for \model}\label{alg:suat}
\end{algorithm*}

\end{document}